\documentclass[11pt]{article}
\usepackage[final]{acl}
\usepackage{caption}
\usepackage{times}
\usepackage{latexsym}
\usepackage{algorithm}
\usepackage{algorithmic}
\usepackage{amsmath,amssymb}
\usepackage{newfloat}
\usepackage{listings}

\usepackage[T1]{fontenc}
\usepackage{booktabs}
\usepackage[utf8]{inputenc}
\usepackage{microtype}
\usepackage{inconsolata}
\usepackage{graphicx}
\title{SOMTab: Set-Order Mamba for Efficient Tabular In-Context Learning}

\author{
  \textbf{Hao Wang\textsuperscript{1}},
  \textbf{Siyu Zhang\textsuperscript{1}},
  \textbf{Wei Ma\textsuperscript{1,*}}
\\
\\
  \textsuperscript{1}Institute of Statistics and Big Data,
  Renmin University of China, Beijing, China
\\
  \small{
    \textsuperscript{*}\textbf{Correspondence:}
    \href{mailto:mawei@ruc.edu.cn}{mawei@ruc.edu.cn}
  }
}

\newcommand{\method}{SOMTab}
\newcommand{\tailmix}{DCH-TailMix}

\newcommand{\LN}{\operatorname{LN}}
\newcommand{\FFN}{\operatorname{FFN}}
\newcommand{\MHA}{\operatorname{MHA}}

\begin{document}
\maketitle

\begin{abstract}
	Tabular foundation models based on in-context learning have recently emerged as strong alternatives to task-specific model fitting. However, the current performance frontier remains dominated by attention-heavy architectures, where attention is used throughout the modeling pipeline. This raises a natural question: is attention necessary at every stage of tabular in-context learning? We introduce SOMTab, a Set-Order Mamba architecture for efficient tabular in-context learning. SOMTab separates representation construction from query-conditioned retrieval. For row and column representations, it maps unordered table tokens into stable latent slots and applies Mamba-based state-space mixing to construct compact representations. For final prediction, it retains attention-based in-context learning to preserve query-conditioned retrieval from labeled context examples. We further introduce DCH-TailMix, a synthetic prior that combines degree-corrected graph heterogeneity with mixed heavy-tailed regimes to diversify synthetic dependency structures. Across tabular benchmarks, SOMTab approaches the performance of strong Transformer-based tabular foundation models while achieving faster inference and lower GPU memory usage, yielding a favorable efficiency--accuracy trade-off.
\end{abstract}

\section{Introduction}

Tabular data are among the most common data formats in scientific and industrial machine learning, with applications in healthcare, finance, biology, materials science, and the social sciences~\citep{borisov2022deep,grinsztajn2022tree}. Despite the remarkable progress of deep learning in vision and language, tabular prediction has long been dominated by tree-based methods such as random forests and gradient-boosted decision trees~\citep{breiman2001random,chen2016xgboost,ke2017lightgbm,prokhorenkova2018catboost}. One key challenge is that tabular datasets are highly heterogeneous: different tables have different feature spaces, feature distributions, sample sizes, class cardinalities, missing-value patterns, and degrees of feature interaction. These properties make it difficult to design a single neural architecture that transfers reliably across datasets~\citep{borisov2022deep,zhu2023xtab,ye2024closer}.

\begin{figure}[t]
	\centering
	\includegraphics[width=1\linewidth]{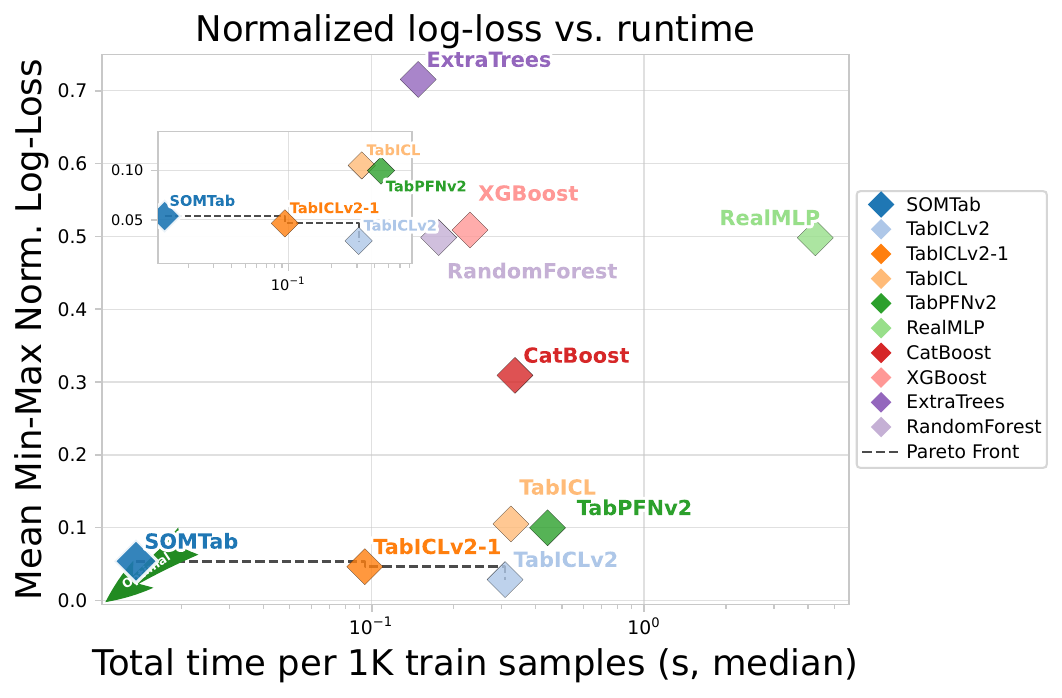}
	\caption{\textbf{Predictive quality--efficiency trade-off on all TALENT classification datasets.}
		The horizontal axis reports the median fit-plus-predict time per 1K labeled context samples, and the vertical axis reports the mean min--max normalized log-loss across datasets. Lower values are better on both axes, and the runtime axis is shown on a logarithmic scale. All methods are rerun on the same H200 server under a common evaluation pipeline; baselines use their official default configurations. The dashed line denotes the empirical Pareto frontier.}
	\label{fig:talent_pareto}
\end{figure}

Tabular foundation models have begun to change this landscape. Built on the framework of Prior-Data Fitted Networks (PFNs), models such as TabPFN, TabPFNv2, TabICL, and TabICLv2 are pretrained on synthetic tabular tasks to amortize prediction across datasets, and apply the learned predictor to a new dataset through in-context learning (ICL), without updating model parameters on the target data~\citep{muller2021transformers,hollmann2022tabpfn,hollmann2025accurate,qu2025tabicl,qu2026tabiclv2}. Given labeled training examples as context and unlabeled test examples as queries, these models are trained to approximate the posterior predictive distribution induced by a synthetic prior over datasets. This paradigm replaces per-dataset model selection and hyperparameter tuning with a single forward-pass predictor, and has achieved strong performance on tabular benchmarks.

\begin{figure*}[t]
	\centering
	\includegraphics[width=0.92\linewidth]{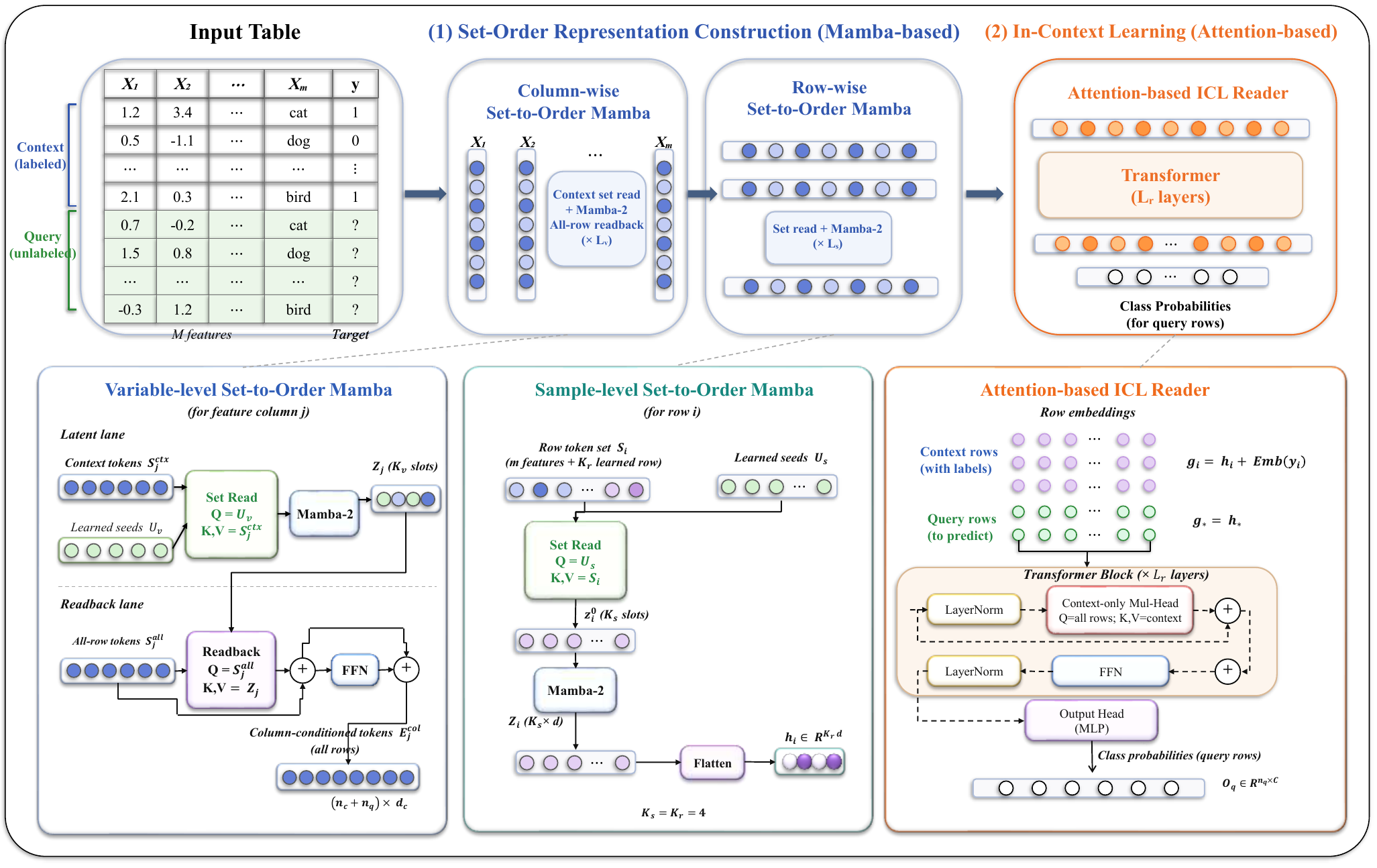}
	\label{fig:SOMTab_architecture}
	\caption{\textbf{Architecture overview of SOMTab.}
		The top illustrates the end-to-end pipeline, where variable- and sample-level Set-to-Order Mamba modules construct compact row representations, followed by an attention-based ICL reader for query prediction. The bottom details the corresponding representation modules and context-only attention mechanism.}
\end{figure*}

However, current leading PFN-style tabular foundation models remain largely built around Transformer attention. Exemplified by TabPFNv2 and its successors, attention serves as a general-purpose table-mixing mechanism: it allows samples, features, and labels to interact so that the model can infer feature distributions, cross-feature dependencies, and task-specific structure from the context~\citep{vaswani2017attention,hollmann2025accurate,qu2025tabicl,qu2026tabiclv2}. This all-to-all comparison is expressive and well aligned with in-context learning, especially in the final prediction stage where test examples must condition on labeled context examples. Yet it also inherits the quadratic cost of attention along the sample or feature dimension, making long-context tabular prediction increasingly expensive as tables grow. This raises a more specific architectural question: must attention be the default operator throughout a tabular foundation model, or is it mainly needed for query-context matching at prediction time? Since earlier table-representation stages often aggregate distributional and interaction information before final prediction, they may be amenable to more efficient linear-complexity sequence operators, such as selective state-space models~\citep{gu2021efficiently,gu2023mamba,dao2024transformers}.

We argue that tabular ICL can be usefully decomposed into two computational roles: representation construction and query-context matching. Representation construction transforms raw cell values and context information into intermediate table or row representations by capturing distributional properties of columns and interactions among features. Query-context matching, in contrast, predicts labels for test examples by conditioning each test representation on labeled examples in the context~\citep{brown2020language,xie2021explanation,muller2021transformers}. Attention is naturally suited for this role because its query-key-value mechanism provides direct query-dependent routing between test and training examples~\citep{vaswani2017attention}. The former role, however, can be viewed as a structured compression problem: the model must aggregate information from unordered rows and columns into useful representations before final prediction. This observation suggests that attention should be allocated selectively rather than treated as the default operator across all stages.

Selective state-space models, exemplified by Mamba, provide a promising alternative for efficient representation construction~\citep{gu2021efficiently,gu2023mamba}. Mamba uses input-dependent state-space dynamics to propagate, suppress, or update information, achieving strong sequence modeling performance while scaling linearly with sequence length~\citep{gu2023mamba}. Yet directly applying Mamba to tabular ICL is nontrivial. Tables are not natural sequences: the order of rows and columns is largely arbitrary~\citep{zaheer2017deep,lee2019set}, whereas Mamba operates through an ordered state-space scan. Moreover, replacing the final ICL stage with pure state-space compression may remove the query-dependent matching behavior that makes attention effective for in-context prediction. Thus, the challenge is not simply to replace Transformers with Mamba, but to integrate state-space mixing into tabular ICL while preserving both the unordered structure of tables and the query-dependent matching capability of attention.

We propose \textbf{SOMTab}, a \textbf{S}et-\textbf{O}rder \textbf{M}amba architecture for efficient tabular in-context learning. SOMTab uses attention selectively in the tabular ICL pipeline. For representation construction, it first maps unordered table tokens into ordered latent slots through a set-to-order mechanism, and then applies Mamba-based mixing over these slots to encode column distributions and row-wise feature interactions efficiently~\citep{zaheer2017deep,lee2019set,gu2023mamba}. For final prediction, SOMTab retains attention-based ICL, allowing test examples to condition on label-relevant information from labeled context examples. This hybrid design reflects our central hypothesis: attention is most valuable for query-context matching, whereas state-space sequence mixing can provide a sufficient and more efficient alternative for table representation once the unordered structure of tabular data is explicitly handled.

In addition to the architecture, we introduce a synthetic dependency prior, \textbf{DCH-TailMix} (\textbf{D}egree-\textbf{C}orrected \textbf{H}eterogeneity with \textbf{Tail} \textbf{Mix}ing), to improve the diversity of pretraining tasks. PFN-style tabular foundation models depend critically on the synthetic prior used during pretraining~\citep{muller2021transformers,hollmann2022tabpfn,hollmann2025accurate,qu2025tabicl,qu2026tabiclv2}. Existing priors based on structural causal models or random graph mechanisms encode useful inductive biases~\citep{pearl2009causality,peters2017elements}, but a single graph-sampling regime may not fully capture the diverse dependency patterns encountered in real-world tables. DCH-TailMix addresses this limitation by combining degree-corrected graph heterogeneity with mixed heavy-tailed regimes, generating diverse dependency structures that expose the model to a broader range of feature-target and feature-feature relationships during pretraining.

Our experiments on TALENT classification datasets show that SOMTab achieves competitive predictive performance among strong Transformer-based tabular foundation models while substantially reducing fit-plus-predict runtime and exhibiting increasingly favorable peak-memory scaling as the labeled context grows. Additional results on TabArena further demonstrate a competitive efficiency--performance trade-off across a broader collection of tabular baselines~\citep{erickson2026tabarena}. Ablations show that both the set-to-order Mamba encoder and the DCH-TailMix prior contribute to the final performance. These results support a simple design principle for tabular ICL: use attention for query--context matching, while employing efficient state-space sequence operators for representation construction when the unordered table structure can be organized into a stable latent order.

Our main contributions are summarized as follows:
\begin{itemize}
	\item We introduce \textbf{SOMTab}, a Set-Order Mamba architecture for tabular in-context learning that organizes unordered table tokens into ordered latent sequences, uses Mamba-based mixing for efficient representation construction, and retains attention for final query-context matching.
	\item We propose \textbf{DCH-TailMix}, a synthetic dependency prior that combines degree-corrected graph heterogeneity with mixed heavy-tailed regimes to generate more diverse feature-target and feature-feature dependency structures for PFN-style pretraining.	
	\item We provide empirical evaluations and ablation studies showing that SOMTab achieves a favorable speed--accuracy trade-off, approaching the predictive performance of strong Transformer-based tabular foundation models while substantially reducing inference time.
\end{itemize}

\section{Related Work}

\paragraph{Tabular foundation models.}
Tabular foundation models aim to apply a pretrained predictor directly to unseen tabular tasks without fitting a separate model from scratch. A central branch is based on Prior-Data Fitted Networks (PFNs), which pretrain neural predictors on tasks sampled from a prior over datasets and perform inference by conditioning on the labeled training set~\citep{muller2021transformers}. TabPFN introduced this paradigm for small tabular classification tasks~\citep{hollmann2022tabpfn}, while TabPFNv2 substantially improved its accuracy, task coverage, and handling of heterogeneous tabular inputs~\citep{hollmann2025accurate}. Subsequent work has improved PFN-style models along several directions, including scalable context construction and localized inference~\citep{thomas2024retrieval}, real-data or hybrid pretraining~\citep{ma2026tabdpt,grinsztajn2025tabpfn}, richer synthetic priors~\citep{zhang2026mitra,zhang2025limix,qu2026tabiclv2}, and more efficient architectures for large tables~\citep{qu2025tabicl,qu2026tabiclv2}. Beyond PFN-style ICL, other tabular foundation models generate task-adaptive predictors through hypernetworks~\citep{bonet2024hyperfast,muller2025mothernet} or leverage column and task semantics through language-model-based representations~\citep{gardner2024large,wen2024supervised}. SOMTab belongs to the PFN-style tabular ICL family, but studies whether state-space mixing can replace attention in representation construction while retaining attention for final query-conditioned retrieval.

\paragraph{Efficient architectures and state-space sequence models.}
The strong performance of PFN-style tabular ICL models is closely tied to attention-based architectures, which enable flexible interactions within and across tabular contexts. However, attention can be costly when used as the default operator throughout the full tabular ICL pipeline. Efficient sequence models, especially structured state-space models, provide an alternative computational primitive for long-context representation~\citep{gu2021efficiently}. 
Mamba introduces input-dependent selective state-space dynamics, allowing the model to propagate, suppress, or update information conditioned on the current token while retaining linear scaling in sequence length~\citep{gu2023mamba}.
Hybrid Mamba--Transformer architectures have also been explored to balance
modeling quality and inference efficiency~\citep{waleffe2024empirical,blakeman2025nemotron}.
Directly applying Mamba to tabular ICL is nontrivial because tables are not natural sequences: rows are exchangeable examples and columns do not have a universal ordering across datasets. Set-based architectures such as Set Transformer address unordered inputs through permutation-invariant or permutation-equivariant attention~\citep{lee2019set}, but still rely on attention as the main representation engine. SOMTab addresses this mismatch by mapping unordered table tokens into ordered latent slots before applying Mamba-based mixing for efficient representation construction.

\paragraph{Synthetic priors for tabular pretraining.}
The inductive bias and performance of PFN-style models are strongly shaped by the synthetic prior used during pretraining. TabPFN uses synthetic priors based on structural causal models and Bayesian neural networks to generate diverse supervised tasks~\citep{hollmann2022tabpfn,pearl2009causality}. Subsequent tabular foundation models enrich these priors with mechanisms such as tree-based generation, heterogeneous feature types, missing values, categorical variables, and more diverse graph and function families~\citep{hollmann2025accurate,qu2025tabicl,qu2026tabiclv2}. These priors implicitly specify the class of learning problems that the model is trained to amortize. SOMTab complements these developments with DCH-TailMix, which combines degree-corrected graph heterogeneity with mixed heavy-tailed regimes.

\section{Method}
\label{sec:method}

\subsection{Tabular In-Context Learning}
\label{sec:tabular_icl}

We focus on supervised tabular classification. A task consists of a labeled context set
\(	D_{\mathrm{train}}=\{(x_i,y_i)\}_{i=1}^{n_{\mathrm{tr}}},\)
where, after preprocessing, each row is represented as $x_i\in\mathbb{R}^{m}$ with $m$ tabular features, and $y_i\in\{1,\ldots,C\}$ denotes its class label. The model is given an unlabeled query set
\(X_{\mathrm{test}}=\{x_j^\ast\}_{j=1}^{n_{\mathrm{te}}},\)
and directly predicts the conditional label distribution
\(q_\theta(y_j^\ast\mid x_j^\ast,D_{\mathrm{train}}),\)
without updating its parameters on the target dataset.

In the PFN-style formulation, synthetic classification tasks are sampled from a prior over datasets and randomly partitioned into context and query examples; the model is then trained by minimizing the negative log-likelihood of the held-out query labels~\citep{muller2021transformers,hollmann2022tabpfn}. During pretraining, the query labels are used only to compute the loss and are never provided as model inputs. For a synthetic task with query pairs
\(D_{\mathrm{test}}=\{(x_j^\ast,y_j^\ast)\}_{j=1}^{n_{\mathrm{te}}},\)
we minimize the query-set cross-entropy

\begin{equation}
\begin{aligned}
	\mathcal{L}_{\mathrm{ICL}}(\theta)
	&=
	\mathbb{E}_{\mathcal{D}\sim p(\mathcal{D}),\,\pi}
	\left[
	\frac{1}{n_{\mathrm{te}}}
	\sum_{j=1}^{n_{\mathrm{te}}}
	\right.\\
	&\qquad\left.
	-
	\log q_\theta
	(y_j^\ast\mid x_j^\ast,D_{\mathrm{train}})
	\right],
\end{aligned}
\label{eq:icl_loss}
\end{equation}
where $\pi$ denotes a random context--query split of the sampled task $\mathcal{D}$. At inference time, the pretrained model is applied to a real dataset by placing the training examples in context and predicting the label distributions of the query examples in a single forward pass.

\subsection{Attention Allocation in Tabular ICL}
\label{sec:attention_allocation}

We organize SOMTab around two computational roles in tabular ICL: representation construction and query-conditioned prediction.
The first jointly maps the task table into row representations,
\begin{equation}
	H
	=
	E_\theta(X;D_{\mathrm{train}})
	=
	[h_1,\ldots,h_{n_{\mathrm{tr}}},
	h_1^\ast,\ldots,h_{n_{\mathrm{te}}}^\ast]^\top,
	\label{eq:row_repr_general}
\end{equation}
where $X=[X_{\mathrm{train}};X_{\mathrm{test}}]$ contains the table values, while only context labels from $D_{\mathrm{train}}$ are available to the representation module.
The encoder summarizes column-wise distributional information and row-wise feature interactions without directly scanning the arbitrary row or column order.

The second role predicts each query label by retrieving information from the labeled context conditioned on its row representation,
\begin{equation}
	q_\theta(y_j^\ast\mid x_j^\ast,D_{\mathrm{train}})
	=
	R_\theta
	\left(
	h_j^\ast,
	\{(h_i,y_i)\}_{i=1}^{n_{\mathrm{tr}}}
	\right).
	\label{eq:context_aggregation_general}
\end{equation}

These roles favor different computational operators.
Representation construction compresses unordered table tokens into fixed-dimensional row embeddings, whereas final prediction requires query-specific access to labeled context examples.
Accordingly, SOMTab uses Set-Order Mamba blocks for representation construction and retains attention in the final row-level ICL reader.
This design reduces attention-heavy computation during table encoding while preserving query-dependent retrieval from the labeled context.

\subsection{Set-to-Order Mamba Block}
\label{sec:set_order_mamba}

Mamba-family selective state-space models efficiently mix ordered sequences, whereas the rows and features of a table have no canonical order. Applying a state-space scan directly to their input order would therefore make the resulting representations depend on arbitrary permutations. Within each representation block, SOMTab instead maps an unordered token collection to a fixed-index latent sequence before applying Mamba-based mixing.

Let
\begin{equation}
	S=[s_1;\ldots;s_N]\in\mathbb{R}^{N\times d},
	\qquad s_i\in\mathbb{R}^{d},
\end{equation}
denote a token collection whose input order is arbitrary. We introduce $K$ learnable seed slots
\begin{equation}
	U=[u_1;\ldots;u_K]\in\mathbb{R}^{K\times d},
\end{equation}
whose fixed indices define the latent sequence order. The seed slots aggregate information from the input set through cross-attention:
\begin{equation}
	\begin{aligned}
		Z'
		&=
		U+
		\mathrm{Attn}\!\big(
		\mathrm{LN}(U),
		\mathrm{LN}(S),
		\mathrm{LN}(S)
		\big),\\
		Z_0
		&=
		Z'+
		\mathrm{FFN}\!\big(
		\mathrm{LN}(Z')
		\big).
	\end{aligned}
	\label{eq:set_to_order}
\end{equation}

Because no positional encoding is attached to the input tokens, permuting the rows of $S$ leaves the latent sequence $Z_0$ unchanged. Its order is instead determined by the seed-slot indices. The latent sequence is then processed by a pre-normalized Mamba-2 residual block~\citep{dao2024transformers}:
\begin{equation}
	\begin{aligned}
		\widetilde{Z}
		&=
		Z_0+
		\operatorname{Mamba2}\!\big(
		\mathrm{LN}(Z_0)
		\big),\\
		Z
		&=
		\widetilde{Z}+
		\mathrm{FFN}\!\big(
		\mathrm{LN}(\widetilde{Z})
		\big).
	\end{aligned}
	\label{eq:mamba_latent}
\end{equation}
The cross-attention in Eq.~\eqref{eq:set_to_order} provides a set-to-slot interface with cost $O(NK)$, while Mamba-2 models interactions along the ordered latent sequence.

When the subsequent stage requires an updated representation for each input token, the latent information is read back to the original tokens:
\begin{equation}
	\begin{aligned}
		\widehat{S}'
		&=
		S+
		\mathrm{Attn}\!\big(
		\mathrm{LN}(S),
		\mathrm{LN}(Z),
		\mathrm{LN}(Z)
		\big),\\
		\widehat{S}
		&=
		\widehat{S}'+
		\mathrm{FFN}\!\big(
		\mathrm{LN}(\widehat{S}')
		\big).
	\end{aligned}
	\label{eq:readback}
\end{equation}

Here, cross-attention transfers the slot-level summary to each input token, while the residual connection preserves its original representation. At the block level, the set-to-order mapping is permutation-invariant, and the readback operation is permutation-equivariant with respect to the input tokens. The latent sequence $Z$ provides a compact set representation, while the optional readback produces updated representations aligned with the original input tokens.

\subsection{SOMTab Architecture}
\label{sec:somtab_architecture}

SOMTab applies the Set-to-Order Mamba block along two complementary axes of a table: across samples within each feature and across features within each sample. The resulting sample representations are then used for row-level in-context prediction.

\paragraph{Representation construction.}
Given a task table $X\in\mathbb{R}^{n\times m}$, where
$n=n_{\mathrm{tr}}+n_{\mathrm{te}}$, we construct a token for each active feature position. Following TabICLv2~\citep{qu2026tabiclv2}, we adopt repeated feature grouping to alleviate feature symmetries. Let $\widetilde{x}_{ij}$ denote the grouped input associated with feature $j$ in row $i$. The initial cell token is
\begin{equation}
	e_{ij}^{(0)}
	=
	\phi_x(\widetilde{x}_{ij})
	+
	\mathbf{1}[i\le n_{\mathrm{tr}}]\psi_y(y_i),
	\qquad
	e_{ij}^{(0)}\in\mathbb{R}^{d}.
	\label{eq:cell_tokenization}
\end{equation}
Thus, label information is injected only into context rows. 

For each feature $j$, we collect its context-row tokens as $S_{j,\mathrm{tr}}^{\mathrm{var}} =\{e_{ij}^{(0)}\}_{i=1}^{n_{\mathrm{tr}}}$.
This set is mapped to latent slots using the set-to-order operation in Eq.~\eqref{eq:set_to_order} and mixed by Mamba as in Eq.~\eqref{eq:mamba_latent}. The resulting feature-level latent representation is then read back, through Eq.~\eqref{eq:readback}, to the tokens of all context and query rows, yielding updated cell representations $\{e_{ij}^{(1)}\}_{i=1}^{n}$. Thus, only context rows determine the latent summary of each feature, while query rows read from this summary without altering it.

For each row $i$, the $m$ updated feature tokens are combined with $K_r=4$ shared learned row tokens to form the unordered sample-level token set
\(
S_i^{\mathrm{sample}}
=
\{r_1,\ldots,r_{K_r},
e_{i1}^{(1)},\ldots,e_{im}^{(1)}\}.
\)
We map this set to $K_s=4$ ordered latent slots and apply Mamba-based mixing. Since $K_s=K_r=4$ in the reported model, the resulting slots are directly concatenated:
\(
h_i
=
\operatorname{Concat}(z_{i1},\ldots,z_{iK_s})
\in\mathbb{R}^{4d}.
\)

\paragraph{Attention-based ICL reader.}
A second label embedding is added to each context row representation, after which the attention-based reader in Eq.~\eqref{eq:context_aggregation_general} predicts the query labels. The early label embedding conditions table representation, whereas the later embedding explicitly provides labeled context pairs to the final reader.

\paragraph{Complexity.}
Let $N=n_{\mathrm{tr}}+n_{\mathrm{te}}$ be the total number of rows, and let $K_v$ and $K_s$ denote the numbers of variable-level and sample-level latent slots. Ignoring hidden-dimension, layer-count, and constant feature-group-size factors, the representation modules cost
\(
O\!\left(m n_{\mathrm{tr}} K_v + m N K_v + N m K_s + mK_v + NK_s\right),
\)
where the first two terms correspond to variable-level aggregation and readback, the third term to sample-level aggregation, and the last two terms to Mamba mixing over latent slots. Since $n_{\mathrm{tr}}\le N$, this simplifies to
\(
O\!\left(Nm(K_v+K_s)\right).
\)
For fixed $K_v$ and $K_s$, representation construction is therefore linear in the number of observed cells, rather than quadratic in the sample or feature dimension as in full attention. The final ICL reader retains row-level attention with cost
\(
O\!\left(n_{\mathrm{tr}}^2+n_{\mathrm{tr}}n_{\mathrm{te}}\right),
\)
which is bounded by $O(N^2)$.

\subsection{DCH-TailMix Synthetic Prior}
\label{sec:dch_tailmix}

We introduce DCH-TailMix, a synthetic prior that generates classification tasks from latent directed acyclic graphs (DAGs) with diverse structural and heavy-tailed dependency regimes.

\paragraph{Degree-corrected graph sampling.}
Let $1,\ldots,L$ denote topologically ordered latent nodes. For each pair $i<j$, an edge $i\rightarrow j$ is sampled with probability
\begin{equation}
	p_{ij}=\sigma(\ell_{ij}),
\end{equation}
where the edge logit combines complementary sources of graph variation:
\begin{equation}
	\ell_{ij}
	=
	\tau+g
	+a_i^{\mathrm{out}}
	+a_j^{\mathrm{in}}
	+b_{c_i,c_j}
	+\lambda\frac{\langle u_i,v_j\rangle}{\sqrt{r}}
	+\epsilon_{ij}.
	\label{eq:dch_logit}
\end{equation}
Here, $\tau$ controls base sparsity, $g$ introduces task-level density variation, and $a_i^{\mathrm{out}}$ and $a_j^{\mathrm{in}}$ create heterogeneous outgoing and incoming degrees. The module-pair bias $b_{c_i,c_j}$ induces block-structured dependencies, while the low-rank term captures correlated connection patterns and $\epsilon_{ij}$ adds edge-level randomness. A separately sampled graph family controls the emphasis placed on these structural components, allowing the prior to generate sparse, modular, dense, correlated, and hub-dominated graphs. Detailed graph-family definitions and sampling probabilities are provided in Appendix B.

\paragraph{Tail-regime mixture.}
TailMix controls the tail behavior of the task-level and node-level graph biases. Rather than relying on a single distribution family, we sample
\begin{equation}
	\rho
	\sim
	\mathrm{Cat}
	(\pi_{\mathrm{base}},\pi_{\mathrm{hier}},\pi_{\mathrm{pareto}}), \\
	(g,a^{\mathrm{out}},a^{\mathrm{in}})
	\sim P_{\rho}.
	\label{eq:tail_regime}
\end{equation}

The independent Student-$t$ regime samples the graph biases independently with controllable tail strength. The hierarchical Student-$t$ regime shares a task-level tail parameter across node biases, producing coherent within-task degree heterogeneity. The symmetric Pareto regime produces occasional extreme positive and negative biases, leading to more pronounced hub-like structures. Thus, DCH controls the structural placement of dependencies, whereas TailMix controls the tail behavior of the biases governing their concentration.

\paragraph{Latent propagation and task construction.}
After sampling the DAG, root nodes are initialized by random vectors, and each non-root node is generated from its parents:
\begin{equation}
	z_j
	=
	f_j\!\left(\{z_i:i\in\mathrm{Pa}(j)\}\right)
	\label{eq:node_propagation}
\end{equation}
where $f_j$ is sampled from function families including smooth, tree-like, thresholding, and interaction-based mappings. Observed features are converted from selected latent nodes, while the target is generated from one to three latent nodes and discretized into class labels. Further details on the tail parameters, nonlinear functions, feature converters, parent truncation, and task-filtering rules are provided in Appendix B.

\subsection{Pretraining Objective}
\label{sec:pretraining_objective}

During pretraining, each mini-batch consists of synthetic classification tasks sampled from DCH-TailMix. For each task, we sample a context--query split, feed the labeled context and unlabeled query rows into SOMTab, and minimize the ICL loss in Eq.~\eqref{eq:icl_loss}. We use a two-stage curriculum over synthetic task sizes and context fractions. Architectural, prior-generation, and training details are provided in Appendices A--C.

\section{Experiments}
We evaluate SOMTab in terms of predictive quality, runtime, and GPU-memory scaling, and compare the attention-based final row-level ICL reader with a Mamba-based alternative.

\subsection{Experimental Setup}

\paragraph{Benchmark and evaluation protocol.}
We evaluate SOMTab on all classification datasets in the TALENT benchmark~\citep{ye2024closer}. For each dataset, we combine the official training and validation splits to form the labeled context and report performance on the official test split. All methods are evaluated using the same data splits and metric computation.

\begin{table}[b]
	\centering
	\small
	\resizebox{\columnwidth}{!}{
		\begin{tabular}{@{}lcccc@{}}
			\toprule
			Method & Norm. LL $\downarrow$ & Accuracy $\uparrow$ & Macro F1 $\uparrow$ & Time/1K $\downarrow$ \\
			\midrule
			SOMTab & 0.054 & 0.849 & 0.785 & 0.019 \\
			TabICLv2 & 0.029 & 0.854 & 0.790 & 0.463 \\
			TabICLv2(1 est.) & 0.046 & 0.853 & 0.789 & 0.151 \\
			TabICL & 0.105 & 0.846 & 0.776 & 0.453 \\
			TabPFNv2 & 0.100 & 0.849 & 0.779 & 0.675 \\
			RealMLP & 0.498 & 0.830 & 0.757 & 4.379 \\
			CatBoost & 0.309 & 0.826 & 0.755 & 0.508 \\
			XGBoost & 0.509 & 0.827 & 0.767 & 0.442 \\
			ExtraTrees & 0.716 & 0.825 & 0.752 & 0.232 \\
			RandomForest & 0.498 & 0.828 & 0.755 & 0.262 \\
			\bottomrule
		\end{tabular}
	}
    \caption{\textbf{Aggregate performance over all TALENT classification datasets.}
		Normalized log-loss, accuracy, and Macro F1 are averaged across datasets, while runtime is the median fit-plus-predict time per 1K labeled context samples. TabICLv2 (default) uses the standard ensemble setting with eight estimators, whereas TabICLv2 (1 estimator) uses a single estimator.}
	\label{tab:main_results}
\end{table}

\begin{figure}[t]
	\centering
	\includegraphics[width=1\linewidth]{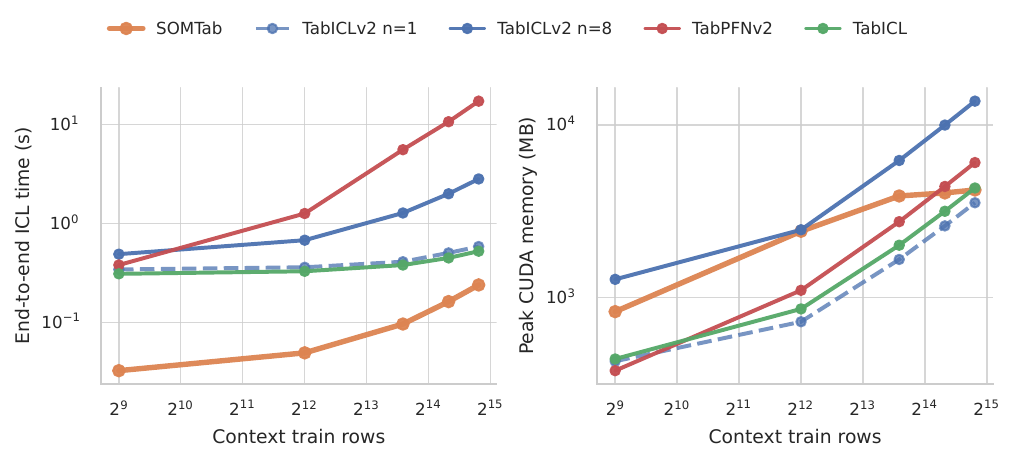}
	\caption{\textbf{Runtime and GPU-memory scaling with context length.}
		We fix $n_{\mathrm{test}}=1024$, $m=32$ features, and $C=10$ classes, and vary the number of context rows from 512 to 28,672. All foundation models use one estimator to isolate architectural scaling from prediction ensembling. Results are measured after one warmup run and report the median of three repeats on identical synthetic data.}
	\label{fig:runtime_memory}
\end{figure}

\paragraph{Baselines and configurations.}
We compare SOMTab with RandomForest~\citep{breiman2001random}, ExtraTrees~\citep{geurts2006extremely}, XGBoost~\citep{chen2016xgboost}, CatBoost~\citep{prokhorenkova2018catboost}, RealMLP~\citep{holzmuller2024better}, TabPFNv2~\citep{hollmann2025accurate}, TabICL~\citep{qu2025tabicl}, and TabICLv2~\citep{qu2026tabiclv2}. All methods are evaluated on the same compute server. For each baseline, we use its official implementation and default configuration, including the default preprocessing and ensemble size.

\paragraph{Metrics and timing.}
Let $L_{m,d}$ denote the test log-loss of method $m$ on dataset $d$. We report the mean min--max normalized log-loss
\begin{equation}
	\widetilde{L}_{m,d}
	=
	\frac{L_{m,d}-L_{\min,d}}
	{L_{\max,d}-L_{\min,d}},
	\qquad
	\overline{L}_m
	=
	\frac{1}{|\mathcal{D}|}
	\sum_{d\in\mathcal{D}}
	\widetilde{L}_{m,d},
	\label{eq:normalized_logloss}
\end{equation}
where $L_{\min,d}$ and $L_{\max,d}$ are the best and worst log-loss values on dataset $d$. Runtime is the median fit-plus-predict time per 1K labeled context samples across datasets. We exclude external preprocessing, model construction, and checkpoint loading, and record peak allocated CUDA memory over the same interval.

\paragraph{SOMTab pretraining.}
The main SOMTab model contains $28.6$M parameters and is trained on synthetic tasks generated by DCH-TailMix. Training follows a two-stage schedule. Stage 1 runs for 300K steps with an initial learning rate of $8\times10^{-4}$ on four 40\,GB NVIDIA A100 GPUs and takes approximately 17 days. Stage 2 continues from the Stage-1 checkpoint for 100K steps with an initial learning rate of $5\times10^{-5}$, requiring approximately six additional days on the same hardware. All training uses float32 precision.

\begin{figure}[t]
	\centering
	\includegraphics[width=1\linewidth]{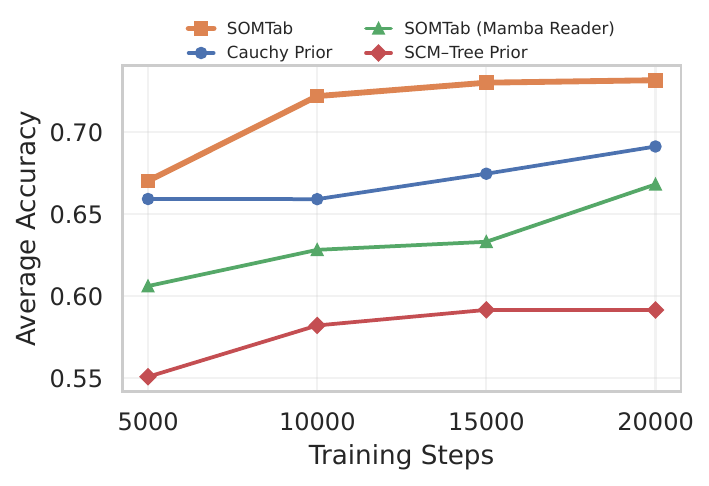}
	\caption{\textbf{Controlled early-training ablation.}
		All variants are trained from scratch for 20K steps under the same configuration. Checkpoints at 5K, 10K, 15K and 20K steps are evaluated on the 15 TALENT datasets where SOMTab has the largest accuracy gaps relative to TabICLv2 in the main benchmark. The vertical axis reports the equally weighted mean accuracy across datasets.}
	\label{fig:ablation}
\end{figure}

\subsection{Predictive Quality--Efficiency Trade-off}

Figure~\ref{fig:talent_pareto} compares probabilistic predictive performance, measured by normalized log-loss, against fit-plus-predict runtime. Its normalized log-loss is close to that of strong tabular foundation models, including TabICLv2 and TabPFNv2, while requiring substantially less runtime and outperforming most tree-based baselines. To separate the effect of prediction ensembling, we additionally report TabICLv2 with a single estimator alongside its official default configuration. Reducing the estimator count lowers the runtime of TabICLv2 but also increases its normalized log-loss, revealing the trade-off introduced by ensembling. Under the matched H200 evaluation, SOMTab requires substantially less runtime than both the default and single-estimator TabICLv2 configurations while maintaining competitive predictive performance.

This result supports our architectural design: by replacing the most computationally expensive attention-based representation construction stages with Mamba-based mixing while retaining attention for the final ICL prediction stage, SOMTab substantially improves efficiency with only a minor impact on predictive quality.

Table~\ref{tab:main_results} provides the aggregate values underlying Figure~\ref{fig:talent_pareto}. SOMTab achieves competitive average
classification performance while maintaining one of the lowest runtimes among high-performing methods. Per-dataset accuracy results on 20 selected TALENT datasets are provided in Appendix E.

We additionally place SOMTab in the broader TabArena benchmark landscape. Appendix F reports an improvability--runtime Pareto comparison and a pairwise win-rate matrix against a broad collection of tabular models. Because the public TabArena baselines retain their original H100 configurations while SOMTab is measured on A100 and H200 GPUs, the runtime comparison provides contextual positioning rather than a hardware-matched speed ranking.

\subsection{Runtime and Memory Scaling}

Figure~\ref{fig:runtime_memory} compares SOMTab with TabICL, TabICLv2, and TabPFNv2 as the labeled context size increases. We fix the numbers of query samples, features, and classes to 1024, 32, and 10, respectively, and vary the number of context rows from 512 to 28,672. For a fair comparison, all models are first evaluated with a single estimator. We additionally report TabICLv2 with its default eight-estimator configuration to quantify the overhead introduced by prediction ensembling.

Runtime and memory are measured under the same evaluation protocol described above. SOMTab exhibits the lowest runtime across the evaluated context lengths, with an increasing advantage as the context size grows. For memory usage, SOMTab can be slightly higher than attention-based baselines in small-context settings due to its latent representation modules; however, as the context size increases, the quadratic memory growth of attention-based architectures becomes dominant. Overall, replacing attention in column- and row-level representation construction with Set-Order Mamba improves long-context efficiency while retaining attention for final in-context prediction.

\subsection{Ablation Studies}

Figure~\ref{fig:ablation} compares the effects of different synthetic priors and final ICL readers. To better reveal differences between variants, we evaluate them on 15 TALENT datasets where SOMTab performs relatively less favorably than TabICLv2 in the main benchmark. The same subset is used for all variants and checkpoints.

The three prior variants share the same SOMTab architecture and differ only in the synthetic prior. They respectively use DCH-TailMix, a random Cauchy graph prior following the TabICLv2~\citep{qu2026tabiclv2} generation mechanism, and the TabICL-style mixture of 70\% MLP-based SCM tasks and 30\% tree-based SCM tasks. DCH-TailMix achieves the highest mean accuracy at the evaluated checkpoints, indicating better transfer under the same pretraining budget.

The fourth variant retains DCH-TailMix and the same Set-Order Mamba representation encoders but replaces the final attention-based ICL reader with a Mamba-based reader. Its lower accuracy under the same training configuration suggests that Mamba-based mixing is effective for table representation construction, whereas attention remains better suited to the final in-context prediction stage.

\section{Conclusion}

In this paper, we revisit the role of attention in tabular in-context learning and introduce SOMTab, a Set-Order Mamba architecture that separates table representation construction from query-conditioned retrieval. By organizing unordered table tokens into stable latent sequences and applying Mamba-based mixing for representation learning while retaining attention for final ICL prediction, SOMTab achieves a more efficient architecture for tabular foundation models. We further introduce DCH-TailMix, a synthetic prior that improves the diversity of dependency structures during pretraining. Extensive experiments show that SOMTab approaches the predictive performance of strong Transformer-based tabular foundation models while substantially reducing inference cost. These results suggest that future tabular foundation models may benefit from assigning different sequence modeling mechanisms to different stages of in-context learning.

\bibliography{custom}

\appendix

\section{Architecture and Implementation Details}
\label{app:model}

We provide the architectural details used in the experiments, including tensor shapes, information flow, and model hyperparameters. We use the notation of the main paper.

\subsection{Task Layout and Cell Tokenization}
\label{app:task_layout}

Let a classification task contain $n_{\mathrm{tr}}$ labeled context rows and
$n_{\mathrm{te}}$ unlabeled query rows, with
$n=n_{\mathrm{tr}}+n_{\mathrm{te}}$. After numerical preprocessing, the model
receives
\begin{equation}
    X=[X_{\mathrm{tr}};X_{\mathrm{te}}]
    \in\mathbb{R}^{n\times m},
    \qquad
    y_{\mathrm{tr}}\in\{0,\ldots,C-1\}^{n_{\mathrm{tr}}},
\end{equation}
where, during pretraining, tasks are sampled with $m\leq 100$ active features and $C\leq 10$ classes. Query labels are used only to compute the pretraining loss and are not provided to the model.

Following TabICLv2~\citep{qu2026tabiclv2}, the input projection uses circular feature grouping with offsets $(0,1,3)$:
\begin{equation}
    \begin{aligned}
        \widetilde{x}_{ij}
        &=
        \left[
            x_{i,j},
            x_{i,\langle j+1\rangle_m},
            x_{i,\langle j+3\rangle_m}
        \right],\\
        \langle r\rangle_m
        &=1+((r-1)\bmod m).
    \end{aligned}
    \label{eq:grouping}
\end{equation}
Grouping is applied to active features before padding. A shared linear projection produces the initial cell token,
\begin{equation}
    e_{ij}^{(0)}
    =
    W_x\widetilde{x}_{ij}+b_x
    +\mathbf{1}[i\leq n_{\mathrm{tr}}]\,\psi_{\mathrm{early}}(y_i),
    \qquad
    e_{ij}^{(0)}\in\mathbb{R}^{d}.
    \label{eq:initial_token}
\end{equation}
Only context rows receive label embeddings. Padded feature positions are masked throughout the sample-level encoder.

\subsection{Set-to-Order Mamba Block}
\label{app:set_to_order}

For an unordered token set $S\in\mathbb{R}^{N\times d}$, let $U\in\mathbb{R}^{K\times d}$ denote the seed vectors obtained by adding an MLP embedding of each normalized slot coordinate to its learned seed vector. Their fixed indices define the latent order, while the input tokens carry no positional encoding. The set read is
\begin{align}
	Z'
	&=
	U+\MHA\!\left(
	\LN(U),\LN(S),\LN(S)
	\right),\\
	Z^{(0)}
	&=
	Z'+\FFN\!\left(
	\LN(Z')
	\right).
	\label{eq:set_read_impl}
\end{align}

The seed slots aggregate information from the input set through cross-attention. Since the input tokens carry no positional information, Eq.~\eqref{eq:set_read_impl} is invariant to permutations of $S$.

The ordered latent slots are mixed by a pre-normalized Mamba-2 residual block~\citep{dao2024transformers},
\begin{align}
    \widetilde{Z}
    &=Z^{(0)}+\operatorname{Mamba2}\!\left(\LN(Z^{(0)})\right),\\
    Z
    &=\widetilde{Z}+\FFN\!\left(\LN(\widetilde{Z})\right).
    \label{eq:mamba_impl}
\end{align}
When token-aligned outputs are needed, the latent sequence is read back to the
original set:
\begin{align}
    \widehat{S}'
    &=S+\MHA\!\left(\LN(S),\LN(Z),\LN(Z)\right),\\
    \widehat{S}
    &=\widehat{S}'+\FFN\!\left(\LN(\widehat{S}')\right).
    \label{eq:readback_impl}
\end{align}
The readback is permutation equivariant with respect to $S$. Cross-attention provides set-to-slot aggregation and readback, while Mamba-2 mixes the ordered latent slots.

\subsection{Variable- and Sample-Level Encoders}
\label{app:representation_encoders}

For each feature $j$, the variable-level encoder reads only the context-row tokens:
\begin{equation}
    S_{j}^{\mathrm{ctx}}
    =
    \{e_{ij}^{(0)}\}_{i=1}^{n_{\mathrm{tr}}}.
\end{equation}

The resulting $K_v$ latent slots are processed by one Mamba-2 residual block and read back to the corresponding tokens of all rows. This set-read--Mamba-2--readback operation constitutes one
variable-level block. \method{} applies $L_v=3$ such blocks sequentially, with the output tokens of one block used as the input tokens of the next.

At the sample level, the $m$ updated feature tokens of each row $i$ are augmented with $K_r=4$ shared learned row-summary tokens. 

The resulting $m+K_r$ tokens are first mapped to $K_s=4$ ordered latent slots. The slots are then processed sequentially by $L_s=3$ Mamba-2 residual blocks, with each block receiving
the output of the preceding block. The final slots are denoted by $Z_i^{\mathrm{sample}}\in\mathbb{R}^{4\times d}$.

The row representation is obtained by flattening these slots:
\begin{equation}
	h_i
	=
	\operatorname{vec}\!\left(Z_i^{\mathrm{sample}}\right)
	\in\mathbb{R}^{4d}.
	\label{eq:row_representation}
\end{equation}

Table~\ref{tab:tensor_shapes} summarizes the main intermediate tensors.
\begin{table}[t]
    \centering
    \small
    \resizebox{\columnwidth}{!}{%
    \begin{tabular}{lll}
        \toprule
        Symbol & Shape & Description \\
        \midrule
        $E^{(0)}$ & $n\times m\times d$ & initial cell tokens \\
		$Z_j^{\mathrm{var}}$ & $K_v\times d$ & latent memory for feature $j$ \\
		$E^{\mathrm{var}}$ & $n\times m\times d$ & context-conditioned cell tokens \\
		$Z_i^{\mathrm{sample}}$ & $4\times d$ & ordered slots for row $i$ \\
		$h_i$ & $4d$ & row representation \\
		$H$ & $n\times 4d$ & representations of all rows \\
		$O_{\mathrm{te}}$ & $n_{\mathrm{te}}\times C$ & query logits \\
        \bottomrule
    \end{tabular}}
    \caption{\textbf{Principal tensors in the \method{} forward pass.}
   Batch dimensions are omitted. The main model uses $d=128$, $K_v=64$, and $K_r=K_s=4$.}
    \label{tab:tensor_shapes}
\end{table}

\subsection{Context-Only ICL Reader}
\label{app:icl_reader}

Before the final reader, \method{} adds a second label embedding to the context row representations:
\begin{equation}
	g_i=
	\begin{cases}
		h_i+\psi_{\mathrm{late}}(y_i), & i\leq n_{\mathrm{tr}},\\
		h_i, & i>n_{\mathrm{tr}}.
	\end{cases}
	\label{eq:late_label}
\end{equation}
Let $G=[g_1;\ldots;g_n]$ and
$G_{\mathrm{ctx}}=G_{1:n_{\mathrm{tr}}}$. Each reader block applies
\begin{align}
	\widetilde{G}
	&=G+\MHA\!\left(
	\LN(G),\LN(G_{\mathrm{ctx}}),\LN(G_{\mathrm{ctx}})
	\right),\\
	G'
	&=\widetilde{G}+\FFN\!\left(\LN(\widetilde{G})\right).
	\label{eq:context_only_reader}
\end{align}
All rows act as queries, while keys and values are restricted to the labeled context. Context rows interact with one another, and each query row attends independently to the context. After 12 reader blocks, a two-layer MLP maps the query representations to class logits.

\section{DCH-TailMix Synthetic Prior}
\label{app:dch_tailmix}

\tailmix{} defines a prior over synthetic classification tasks through degree-corrected DAGs, mixed tail regimes, and heterogeneous nonlinear mechanisms. This section gives the sampling distributions, observation model,
and task-validity criteria.

\subsection{Generation Procedure}
\label{app:prior_pipeline}

Algorithm~\ref{alg:tailmix} summarizes the generation procedure. Sequence length, feature count, class count, and context size are determined by the training curriculum. Within each task, related scalar and categorical
hyperparameters share task-specific distributions. For a scalar parameter, we
first draw
\begin{equation}
    t\sim\mathcal{U}(0,1),
    \qquad
    s\sim\operatorname{LogUniform}(0.1,10^4),
\end{equation}
use $(st,s(1-t))$ as beta-distribution parameters, and map the resulting draw to the required linear or logarithmic interval. For categorical parameters, task-specific probabilities are obtained by normalizing log-normal weights whose log-standard deviation is sampled from $[0.2,2.0]$. Sharing these distributions within a task induces correlated generator settings.

\begin{algorithm}[t]
\caption{DCH-TailMix task generation}
\label{alg:tailmix}
\begin{algorithmic}[1]
\REQUIRE sequence length $n$, feature count $m$, number of classes $C$
\STATE Sample $L\in\{2,\ldots,32\}$ latent nodes on a logarithmic scale
\STATE Sample a graph family and a tail regime
\STATE Sample a topologically ordered DCH DAG and truncate parent sets
\STATE Sample root states and propagate non-root latent variables
\STATE Select one to three target nodes and eligible observed-feature nodes
\STATE Convert selected latent states to numerical or categorical columns
\STATE Construct and standardize a continuous target signal
\STATE Convert the target signal to class labels and permute feature and label identities
\STATE Reject degenerate tasks or invalid context--query splits
\RETURN a synthetic classification task $(X,y)$
\end{algorithmic}
\end{algorithm}

\subsection{Degree-Corrected Graph Sampling}
\label{app:dch_graph}

For each topologically valid pair $i<j$, the edge $i\rightarrow j$ is sampled as
\begin{equation}
    A_{ij}\sim\operatorname{Bernoulli}(\sigma(\ell_{ij})),
\end{equation}
where
\begin{equation}
    \ell_{ij}
    =
    \tau+g+a_i^{\mathrm{out}}+a_j^{\mathrm{in}}
    +b_{c_i,c_j}
    +\lambda\frac{\langle u_i,v_j\rangle}{\sqrt{r}}
    +\epsilon_{ij}.
    \label{eq:dch_logit}
\end{equation}
The global, outgoing, and incoming scales are sampled log-uniformly from
$[0.2,1.5]$, $[0.3,2.0]$, and $[0.3,2.0]$, respectively. The edge-level noise satisfies
$\epsilon_{ij}\mid s_{\epsilon}\sim\mathcal{N}(0,s_{\epsilon}^{2})$,
where $s_{\epsilon}$ is sampled log-uniformly from $[0.05,0.8]$. Before Bernoulli sampling, logits are clipped
to $[-6,6]$ and probabilities to $[0.002,0.995]$.

When the modular component is active, the number of modules is sampled uniformly from $\{2,\ldots,5\}$. The module-pair bias includes a Gaussian term scaled by a draw from $[0.2,1.0]$ and a within-module diagonal bias sampled from $[0.5,2.5]$. The low-rank dimension is sampled uniformly from $\{2,\ldots,\min(8,L)\}$, with weight sampled log-uniformly from $[0.1,1.5]$.

\begin{table*}[t]
    \centering
    \small
    \resizebox{\textwidth}{!}{%
    \begin{tabular}{lclll}
        \toprule
        Graph family & Probability & Base sparsity $\tau$ & Structural terms & Parent cap \\
        \midrule
        controlled heavy-tail & $0.40$ & $\mathcal{U}(-4.0,-0.5)$
            & degree corrections; $\lambda=0$ & at most $8$ \\
        modular heavy-tail & $0.25$ & $\mathcal{U}(-4.0,-0.5)$
            & module-pair and low-rank biases & at most $8$ \\
        low-rank correlated connectivity & $0.15$ & $\mathcal{U}(-4.0,-0.5)$
            & correlated low-rank edge logits & at most $8$ \\
        dense weak-dependency & $0.20$ & $\mathcal{U}(-2.4,-0.8)$
            & denser modular and low-rank effects & at most $12$ \\
        \bottomrule
    \end{tabular}}
    \caption{\textbf{Graph-family mixture in \tailmix{}.}
    For dense weak-dependency graphs, the parent cap for node $j$ is at least $\min(8,j-1)$ under one-based topological indexing. When the initial Bernoulli sample for the final node is too sparse, additional
    high-probability candidate edges are inserted before applying this cap.}
    \label{tab:graph_mixture}
\end{table*}

\begin{figure*}[t]
    \centering
    \includegraphics[width=0.96\textwidth]{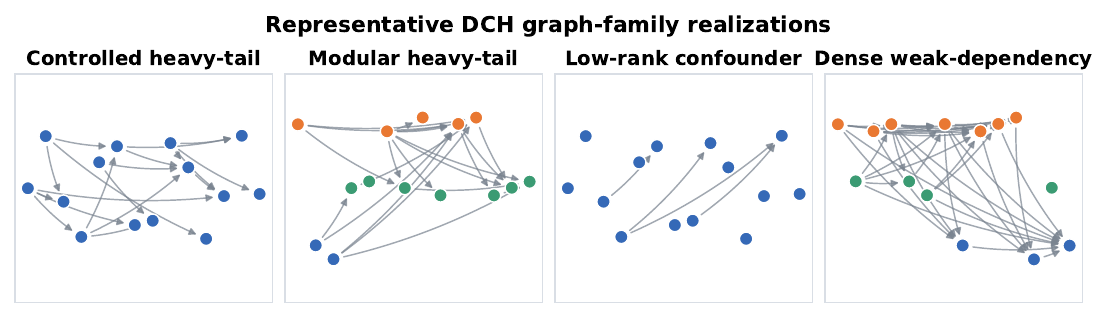}
    \caption{\textbf{Representative realizations from the four DCH graph families.} Nodes are arranged in topological order from left to right; colors identify sampled modules when modular structure is active. Each panel shows one sampled graph.}
    \label{fig:dch_graphs}
\end{figure*}

Table~\ref{tab:graph_mixture} lists the graph-family mixture, and Figure~\ref{fig:dch_graphs} shows representative samples. After edge sampling, parent sets are truncated to bound the input dimension of the random node functions. Observed features are drawn with replacement from a latent pool that includes nodes sharing ancestry with the target, allowing correlated and occasionally duplicated columns.

\subsection{Tail-Regime Mixture}
\label{app:tail_regimes}

The tail regime is sampled once per task:
\begin{equation}
    \rho\sim\operatorname{Cat}(0.70,0.20,0.10),
    \qquad
    (g,a^{\mathrm{out}},a^{\mathrm{in}})\sim P_{\rho}.
\end{equation}
The mixture contains three regimes:
\begin{enumerate}
    \item \textbf{Independent Student-$t$ (0.70).}
    The three degrees of freedom are sampled independently as
    $\nu_g,\nu_{\mathrm{out}},\nu_{\mathrm{in}}\sim\mathcal{U}(1,10)$.
    \item \textbf{Hierarchical Student-$t$ (0.20).}
    A shared base degree of freedom is sampled from
    \begin{equation}
        \nu_0\sim
        \begin{cases}
            \mathcal{U}(1,2), & 0.35,\\
            \mathcal{U}(2,5), & 0.40,\\
            \mathcal{U}(5,10), & 0.25.
        \end{cases}
    \end{equation}
    Each realized value is
    $\nu_q=\operatorname{clip}(\nu_0e^{\delta_q},1,10)$ with
    $\delta_q\sim\mathcal{N}(0,0.15^2)$ and
    $q\in\{g,\mathrm{out},\mathrm{in}\}$.
    \item \textbf{Symmetric Pareto (0.10).}
    With $\alpha\sim\mathcal{U}(1.5,5.0)$,
    $U\sim\mathcal{U}(0,1)$, and an equiprobable sign
    $s\in\{-1,+1\}$, a raw draw is
    \begin{equation}
        r=s\left(U^{-1/\alpha}-1\right).
    \end{equation}
    Vector-valued draws are normalized by the $0.75$ quantile of $|r|$ before applying their sampled scale.
\end{enumerate} 

\begin{figure}[t]
    \centering
    \includegraphics[width=\columnwidth]{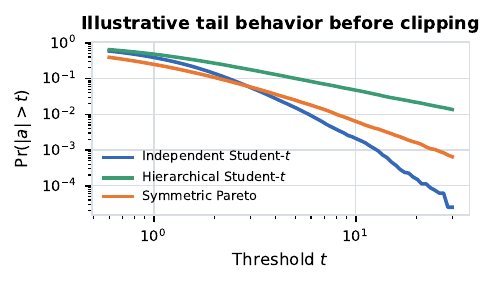}
    \caption{\textbf{Illustrative tail behavior of the three mixture
    components before clipping.} Curves show representative draws of the
    regime parameters.}
    \label{fig:tail_survival}
\end{figure}

In every regime, $g$ is clipped to $[-5,5]$, and the incoming and outgoing corrections are clipped to $[-6,6]$. Figure~\ref{fig:tail_survival} shows representative survival curves before clipping.

\subsection{Latent States and Random Functions}
\label{app:random_functions}

The dimension of each latent node is sampled logarithmically from 1 to 32. Root states are drawn from one of four families: independent Gaussian, uniform on $[-1,1]^d$, uniform in the unit ball, or a random-covariance model. With probability $0.35$, a random linear map and activation are applied to the root state.

For a non-root node with parent states $\{z_p:p\in\operatorname{Pa}(j)\}$, the parents are either concatenated and passed through a single function, or processed separately and combined by a normalized sum, product, maximum, or log-sum-exp. The single-function form is used for one parent and is sampled with probability $0.5$ otherwise. Table~\ref{tab:function_library} lists the function families. Their probabilities are sampled at the task level rather than fixed to a uniform distribution.

\begin{table*}[t]
    \centering
    \small
    \resizebox{\textwidth}{!}{%
    \begin{tabular}{lll}
        \toprule
        Family & Construction & Sampled ranges / constraints \\
        \midrule
        linear & random matrix projection
            & Gaussian, low-rank, SVD-shaped, kernel, or activated matrix; input dimension $\leq128$ \\
        neural & shallow random MLP
            & depth $1$--$3$; width $1$--$127$; sampled input/output activations \\
        tree & ensemble of random split trees
            & $1$--$128$ trees; depth $1$--$7$; input dimension $\leq32$ \\
        discretization & nearest-center partition followed by projection
            & $2$--$255$ centers; $\ell_p$ distance with $p\in[0.5,4]$ log-uniform \\
        GP-like & finite random Fourier-feature map
            & $256$ features; axis-aligned probability $0.5$; spectral tail $a\in[2,20]$ log-uniform \\
        quadratic & symmetric quadratic form with a bias coordinate
            & input dimension $\leq20$ \\
        EM-like & mixture-responsibility transform followed by projection
            & at least $2$ components; $p\in[1,4]$; response power in $[1,2]$ \\
        product & standardized product of two sampled subfunctions
            & subfunctions exclude neural, EM-like, and product recursion \\
        \bottomrule
    \end{tabular}}
    \caption{\textbf{Random function library for latent propagation.}}
    \label{tab:function_library}
\end{table*}

The activation library comprises GELU, ReLU, $\tanh$, sine, cosine, sigmoid, and absolute value. After propagation, node states are standardized, dimension-wise reweighted, normalized by their mean sample norm, and rescaled by a factor sampled log-uniformly from $[0.1,10]$.

\begin{figure*}[t]
    \centering
    \includegraphics[width=0.96\textwidth]{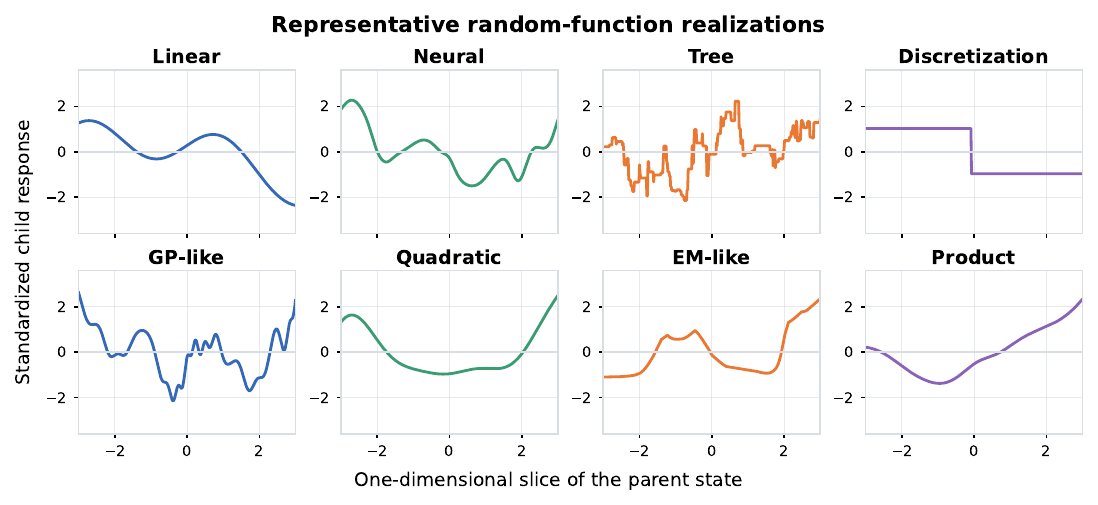}
    \caption{\textbf{Representative one-dimensional slices of the random
    function families.} Each panel shows the response of one function family
    along a fixed low-dimensional parent trajectory, illustrating smooth,
    thresholded, piecewise, and interaction-based mechanisms.}
    \label{fig:random_functions}
\end{figure*}

\subsection{Observation and Classification Adapters}
\label{app:observation_adapter}

The observation model samples a task-level categorical rate
\begin{equation}
    q_{\mathrm{cat}}
    =
    \operatorname{clip}\!\left(
        \mathcal{U}(-0.5,1.2),0,1
    \right)
\end{equation}
and converts each selected latent feature to categorical form independently with probability $q_{\mathrm{cat}}$. Numerical features are one-dimensional latent projections. With probability $0.35$, a numerical feature is min--max scaled and transformed by a Kumaraswamy distribution whose two shape parameters are sampled log-uniformly from $[0.2,5]$.

Categorical features have cardinality between 2 and 9 and use one of two converters. The nearest-center converter uses at most 12 latent dimensions and an $\ell_p$ distance with log-uniform $p\in[0.5,4]$. The softmax converter uses a standardized random projection, category biases, and a separation parameter sampled log-uniformly from $[0.1,10]$. An additional numerical-to-categorical conversion is applied with task-level probability $0.2$, followed by feature standardization, outlier clipping, random feature permutation, and zero padding.

One to three target nodes are combined by the same random function mechanism to produce a standardized continuous target. The class count is set to $C=2$ with probability $1/2$ and is otherwise sampled uniformly from
$\{2,\ldots,10\}$. The resulting marginal probability is $5/9$ for binary tasks and $1/18$ for each class count from 3 to 10. The continuous target is discretized using either randomly selected observed-value thresholds or
Gaussian thresholds, and the class identities are then permuted.

\subsection{Validity Conditions}
\label{app:validity}

Degenerate tasks are removed using validity checks that do not require fitting a separate predictor. A task is rejected if the features or target contain non-finite values, if fewer than $\max(1,\min(2,m))$ features have
non-negligible variance, or if the largest absolute feature--target marginal correlation is at most $0.015$ on a subsample of at most 2048 rows. Constant columns are removed after classification conversion. The row order is then resampled up to ten times until the context and query partitions contain the same set of at least two classes.

\section{Pretraining Details}
\label{app:training}

\subsection{Two-Stage Curriculum}
\label{app:curriculum}

The main model is trained entirely on \tailmix{} tasks. Stage~1 covers a broad range of task sizes and context fractions. Stage~2 continues from the Stage-1 checkpoint with a lower learning rate. Table~\ref{tab:training_schedule} summarizes the curriculum.

\begin{table}[t]
    \centering
    \small
    \resizebox{\columnwidth}{!}{%
    \begin{tabular}{lcc}
        \toprule
        Setting & Stage 1 & Stage 2 \\
        \midrule
        Initialization & random & Stage-1 checkpoint \\
        Training steps & $300{,}000$ & $100{,}000$ \\
        Sequence length & $400$--$8192$ & $400$--$8192$ \\
        Length sampling & log-uniform & log-uniform \\
        Context fraction & $\mathcal{U}(0.1,0.9)$ & $\mathcal{U}(0.5,0.9)$ \\
        Initial learning rate & $8\times10^{-4}$ & $5\times10^{-5}$ \\
        Scheduler & cosine + warmup & polynomial decay \\
        Warmup & $2\%$ & none \\
        Final learning rate & approximately $0$ & $5\times10^{-6}$ \\
        Global / micro batch & $64$ / $1$ & $64$ / $1$ \\
        Numerical precision & float32 & float32 \\
        \bottomrule
    \end{tabular}}
    \caption{\textbf{Two-stage pretraining curriculum.}
    Sequence length is the total number of context and query rows.}
    \label{tab:training_schedule}
\end{table}

Stage~2 uses polynomial decay with power two. Each synthetic task contains between 2 and 100 features and at most ten classes, with the class-count distribution specified in Appendix~\ref{app:observation_adapter}. NumPy and
PyTorch use random seed 42 in all training runs.

\subsection{Optimization and Compute}
\label{app:optimization}

Training uses a hybrid Muon--AdamW optimizer. Muon is applied to non-embedding matrix parameters with momentum $0.95$ and five Newton--Schulz iterations. Embedding and lower-dimensional parameters use AdamW with
$(\beta_1,\beta_2)=(0.9,0.95)$ and $\epsilon=10^{-8}$. Weight decay is set to zero, and the global gradient norm is clipped at 1.0.

Training is distributed over four NVIDIA A100 GPUs with 40\,GB of memory per GPU. Each global batch contains 64 independently generated tasks. The per-device micro-batch size is one, and gradient accumulation yields the global
batch size. Stage~1 requires approximately 17 days, followed by approximately six days for Stage~2 on the same hardware. The final Stage-2 checkpoint is used for evaluation.

\section{Additional Evaluation Details and Results}
\label{app:evaluation}

\subsection{TALENT Metrics and Aggregation}
\label{app:talent_metrics}

For each TALENT classification dataset~\citep{ye2024closer}, the official training and validation splits form the labeled context, and the official test split forms the query set. Accuracy, Macro F1, and log-loss are computed on the test rows, with equal weight assigned to each dataset in aggregate results.

For model $a$ and dataset $d$, log-loss is normalized across the methods in the common result table:
\begin{equation}
    \widetilde{L}_{a,d}
    =
    \frac{L_{a,d}-\min_b L_{b,d}}
    {\max_b L_{b,d}-\min_b L_{b,d}},
    \qquad
    \overline{L}_a
    =
    \frac{1}{|\mathcal{D}|}
    \sum_{d\in\mathcal{D}}\widetilde{L}_{a,d}.
    \label{eq:normalized_logloss}
\end{equation}
All methods in an aggregate comparison are evaluated on the same intersection of datasets.

\subsection{Timing and Memory Measurement}
\label{app:timing}

For methods rerun on TALENT, timing includes model fitting, when applicable, and prediction. Dataset loading, external preprocessing, model construction, and checkpoint loading are excluded. Runtime is normalized by the number of labeled context rows and reported per 1000 context examples; aggregate runtime is the median across datasets. Peak GPU memory is the maximum allocated CUDA memory during the same interval.

The controlled context-scaling experiment fixes $n_{\mathrm{te}}=1024$, $m=32$, and $C=10$, and varies $n_{\mathrm{tr}}$ from 512 to 28,672. All foundation models use one estimator, with the default eight-estimator
TabICLv2 configuration reported separately. Each point is measured after one warmup run and reports the median of three repetitions on identical synthetic inputs.

\subsection{Controlled Early-Training Ablation}
\label{app:ablation}

The controlled ablation compares synthetic priors and final ICL readers under the same training budget. All variants are initialized randomly, trained for 20K steps, and evaluated every 5K steps. Evaluation uses the same subset of 15 TALENT datasets where SOMTab performs relatively less favorably than TabICLv2 in the main benchmark. This subset is used only for the ablation analysis.

\begin{table}[t]
    \centering
    \small
    \resizebox{\columnwidth}{!}{%
    \begin{tabular}{lll}
        \toprule
        Variant & Representation & Prior / reader change \\
        \midrule
        SOMTab & Set-Order Mamba & DCH-TailMix + attention reader \\
        Cauchy & Set-Order Mamba & random Cauchy graph prior \\
        SCM--Tree & Set-Order Mamba & $70\%$ MLP-SCM + $30\%$ tree-SCM \\
        Tab-Full & Set-Order Mamba & DCH-TailMix + Mamba reader \\
        \bottomrule
    \end{tabular}}
    \caption{\textbf{Variants in the controlled ablation.}
    All unspecified architecture and optimization settings are shared.}
    \label{tab:ablation_variants}
\end{table}

\begin{figure}[t]
    \centering
    \includegraphics[width=\columnwidth]{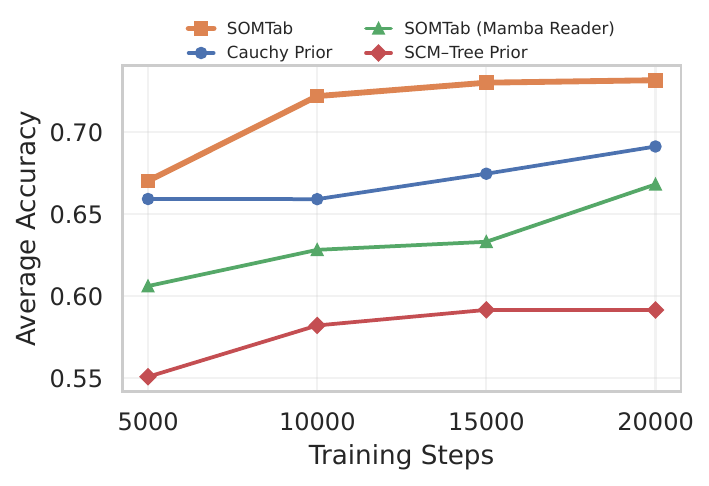}
    \caption{\textbf{Early-training accuracy under a matched 20K-step budget.}
    Values are equally weighted means over the fixed 15-dataset subset.}
    \label{fig:training_curve}
\end{figure}

\begin{table*}[t]
	\centering
	\small
	\setlength{\tabcolsep}{4pt}
	\resizebox{\textwidth}{!}{
		\begin{tabular}{lccccccccc}
			\toprule
			Dataset & SOMTab & TabICLv2 & TabICL & TabPFNv2 &
			RealMLP & CatBoost & XGBoost & ExtraTrees & RandomForest \\
			\midrule
			mice\_protein\_expression
			& 1.000 & 1.000 & 1.000 & 1.000 & 1.000
			& 0.972 & 0.954 & 1.000 & 0.986 \\
			
			shuttle
			& 1.000 & 0.999 & 0.999 & 0.999 & 1.000
			& 1.000 & 1.000 & 0.999 & 1.000 \\
			
			JapaneseVowels
			& 0.997 & 0.999 & 0.999 & 0.999 & 0.990
			& 0.960 & 0.979 & 0.985 & 0.974 \\
			
			gas-drift
			& 0.997 & 0.997 & 0.997 & 0.996 & 0.996
			& 0.991 & 0.995 & 0.996 & 0.994 \\
			
			pendigits
			& 0.996 & 0.997 & 0.997 & 0.996 & 0.998
			& 0.986 & 0.989 & 0.991 & 0.991 \\
			
			\midrule
			thyroid
			& 0.994 & 0.994 & 0.993 & 0.997 & 0.994
			& 0.995 & 0.997 & 0.983 & 0.997 \\
			
			Wilt
			& 0.993 & 0.993 & 0.993 & 0.994 & 0.993
			& 0.988 & 0.985 & 0.981 & 0.982 \\
			
			GAMETES-2Way-20
			& 0.628 & 0.678 & 0.669 & 0.678 & 0.669
			& 0.634 & 0.650 & 0.578 & 0.588 \\
			
			eye\_movements
			& 0.797 & 0.838 & 0.704 & 0.854 & 0.698
			& 0.647 & 0.725 & 0.730 & 0.695 \\
			
			autoUniv-au4-2500
			& 0.704 & 0.724 & 0.516 & 0.706 & 0.488
			& 0.660 & 0.670 & 0.568 & 0.608 \\
			
			eeg-eye-state
			& 0.993 & 0.994 & 0.990 & 0.987 & 0.985
			& 0.894 & 0.930 & 0.953 & 0.930 \\
			
			Indian\_pines
			& 0.915 & 0.957 & 0.939 & 0.939 & 0.922
			& 0.837 & 0.888 & 0.888 & 0.885 \\
			
			autoUniv-au7-1100
			& 0.427 & 0.405 & 0.409 & 0.432 & 0.341
			& 0.382 & 0.395 & 0.373 & 0.418 \\
			
			Firm-Teacher
			& 0.875 & 0.877 & 0.878 & 0.850 & 0.874
			& 0.858 & 0.845 & 0.805 & 0.812 \\
			
			FOREX\_audchf-day-High
			& 0.752 & 0.747 & 0.749 & 0.763 & 0.755
			& 0.649 & 0.662 & 0.665 & 0.668 \\
			
			led24
			& 0.728 & 0.736 & 0.736 & 0.731 & 0.738
			& 0.731 & 0.706 & 0.722 & 0.730 \\
			
			heloc
			& 0.728 & 0.734 & 0.725 & 0.729 & 0.732
			& 0.724 & 0.709 & 0.725 & 0.719 \\
			
			page-blocks
			& 0.978 & 0.977 & 0.979 & 0.980 & 0.970
			& 0.976 & 0.973 & 0.975 & 0.975 \\
			
			waveform-5000
			& 0.867 & 0.867 & 0.865 & 0.866 & 0.864
			& 0.848 & 0.847 & 0.862 & 0.852 \\
			
			abalone
			& 0.644 & 0.641 & 0.642 & 0.652 & 0.638
			& 0.632 & 0.614 & 0.605 & 0.632 \\
			\bottomrule
		\end{tabular}
	}
    	\caption{\textbf{Per-dataset accuracy results on selected TALENT datasets.}
		We report accuracy on 20 TALENT classification datasets. The final 15 rows form the fixed subset used in the controlled early-training ablation.
        	\label{tab:talent_per_dataset_acc}
	}
\end{table*}

At each evaluated checkpoint, \tailmix{} achieves higher mean accuracy than the Cauchy and SCM--Tree priors under the same architecture~\citep{hollmann2025accurate,qu2025tabicl,qu2026tabiclv2}. The Mamba Reader variant, which replaces the final attention reader with Mamba, remains less accurate throughout training. These results support the use of Mamba for representation construction and attention for final in-context retrieval.

\section{Additional Results on TALENT}

\subsection{Per-Dataset Accuracy Results}

Table~\ref{tab:talent_per_dataset_acc} reports per-dataset accuracy on 20 TALENT classification datasets. The final 15 datasets constitute the fixed evaluation subset used in the controlled early-training ablation in
Appendix D.3.

\section{Additional Evaluation on TabArena}
\label{app:tabarena}

\begin{figure}[t]
	\centering
	\includegraphics[width=0.9\linewidth]{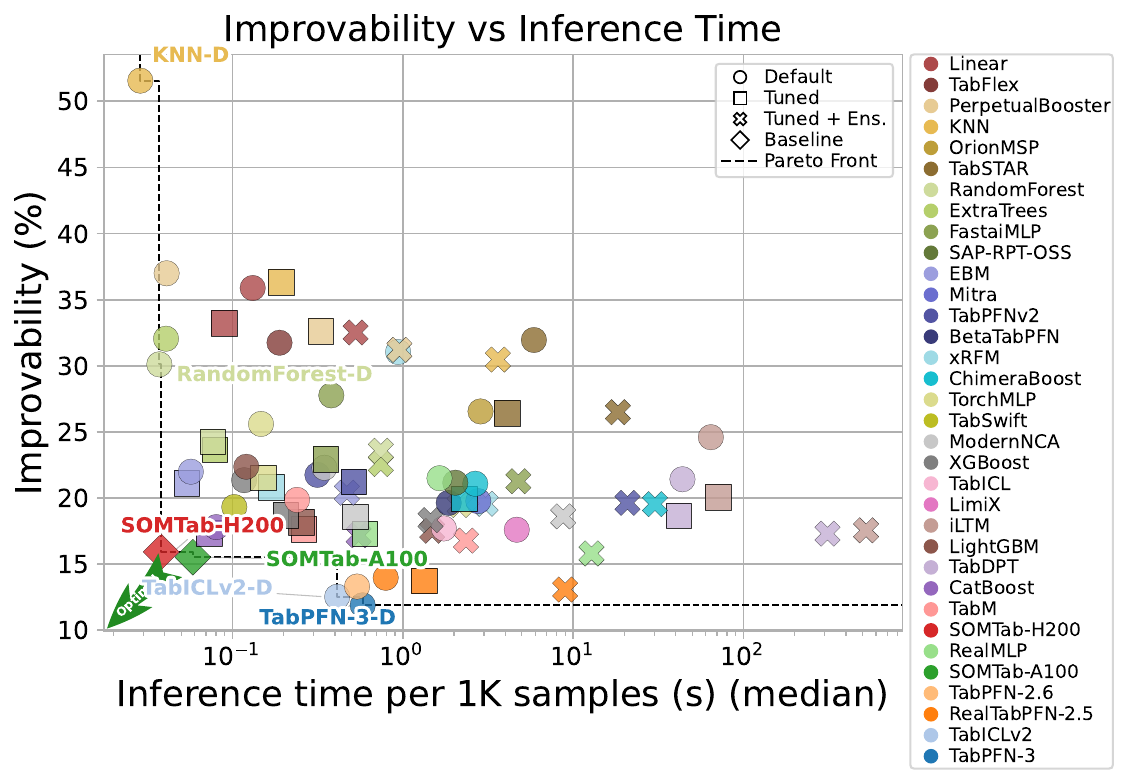}
	\caption{\textbf{Performance--efficiency comparison on TabArena.}
		The x-axis shows inference time per 1000 samples on a logarithmic scale,
		and the y-axis reports the average performance gap to the best method on
		each dataset (lower is better). SOMTab-H200 and SOMTab-A100 denote
		measurements obtained on H200 and A100 GPUs, respectively, while other
		results are taken from the official TabArena evaluation.}
	\label{fig:tabarena}
\end{figure}

Figure~\ref{fig:tabarena} summarizes the performance--efficiency
comparison on the TabArena benchmark~\citep{erickson2026tabarena}.
For each dataset, TabArena reports the performance gap between a model and
the best-performing method on that dataset. We average these gaps across
datasets, where a smaller value indicates that a model is closer to the best
performance.

The runtime results should be interpreted as a reference comparison rather
than a strictly controlled hardware benchmark. The public TabArena results
are reported using their original evaluation environment with H100 GPUs,
while we do not have access to H100 hardware. To reduce the impact of GPU
differences, we additionally evaluate \method{} on both H200 and A100 GPUs.
The two measurements provide a more reliable indication of the practical
efficiency of \method{} across different GPU generations.

\subsection{Evaluation Protocol}
\label{app:tabarena_protocol}

We follow the public TabArena evaluation protocol and include the models
provided by the benchmark. Figure~\ref{fig:tabarena} compares the
performance--efficiency trade-off, while
Figure~\ref{fig:appendix_winrate_matrix} reports pairwise comparisons
between models.

\subsection{Pairwise Win-rate Comparison}
\label{app:tabarena_winrate}

\begin{figure}[t]
	\centering
	\includegraphics[width=0.9\linewidth]{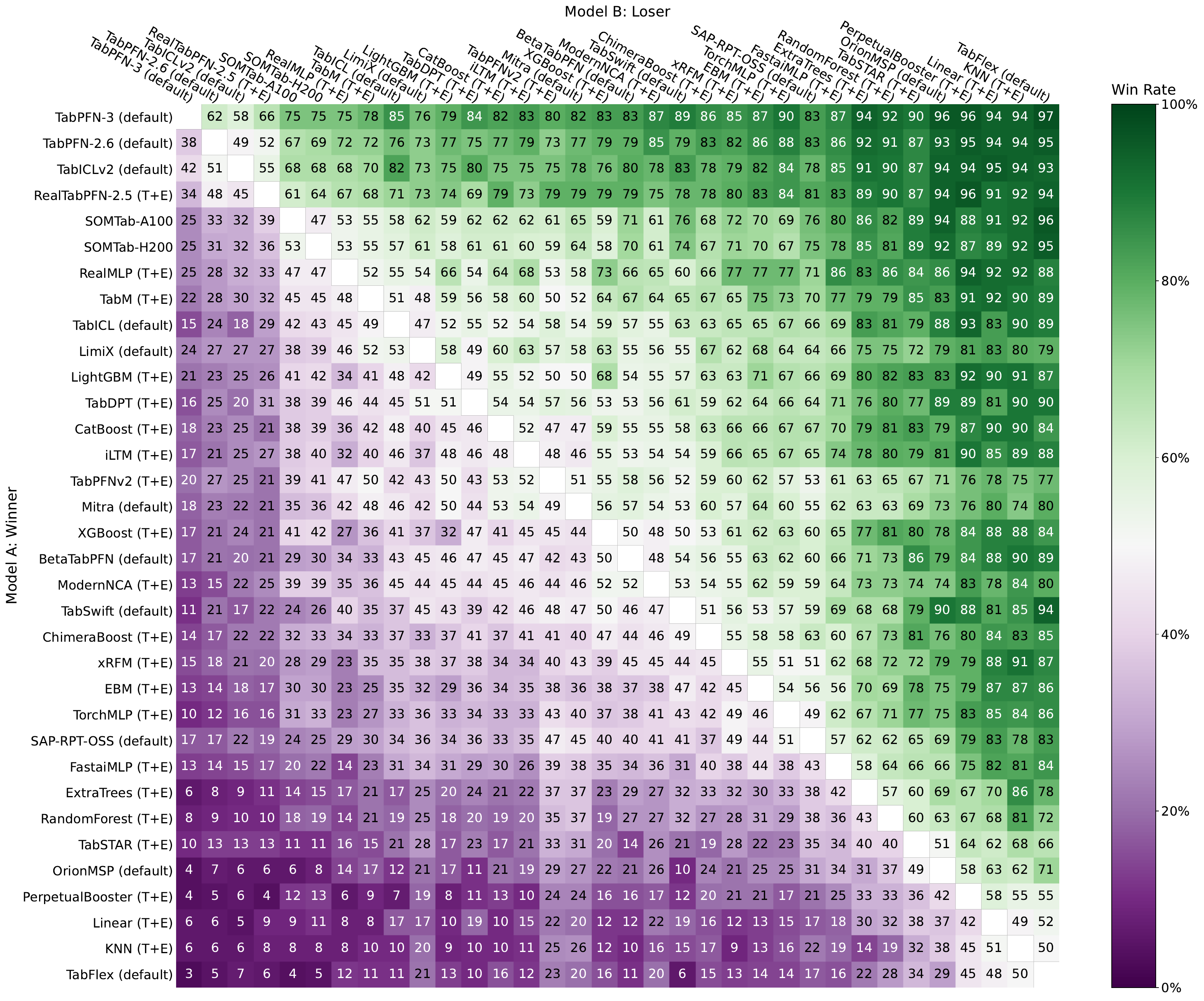}
	\caption{\textbf{Pairwise win-rate comparison on TabArena.}
		Each entry denotes the fraction of datasets where the row model achieves
		higher predictive performance than the column model.}
	\label{fig:appendix_winrate_matrix}
\end{figure}

Figure~\ref{fig:appendix_winrate_matrix} reports pairwise win rates on
TabArena. Each entry shows the fraction of datasets where the model in the
row achieves higher predictive performance than the model in the column.
This comparison reveals how frequently one model outperforms another across
individual datasets rather than relying only on averaged benchmark scores.

\end{document}